\documentclass[letterpaper]{article} 
\usepackage{aaai25}  
\usepackage{times}  
\usepackage{helvet}  
\usepackage{courier}  
\usepackage[hyphens]{url}  
\usepackage{graphicx} 
\usepackage{multirow}
\usepackage{booktabs}
\usepackage{natbib}  
\usepackage{caption} 
\usepackage{algorithm}
\usepackage{algorithmic}
\usepackage{amsmath}
\usepackage{amssymb} 

\usepackage{newfloat}
\usepackage{listings}
\DeclareCaptionStyle{ruled}{labelfont=normalfont,labelsep=colon,strut=off} 
\floatstyle{ruled}
\newfloat{listing}{tb}{lst}{}
\floatname{listing}{Listing}
\usepackage{xcolor}

\title{Hyper Adversarial Tuning for Boosting Adversarial Robustness \\ of 
Pretrained Large Vision Models}
\author{
    Kangtao Lv\textsuperscript{\rm 1}, Huangsen Cao\textsuperscript{\rm 1}, Kainan Tu\textsuperscript{\rm 1,2}, Yihuai Xu\textsuperscript{\rm 1,3}, Zhimeng Zhang\textsuperscript{\rm 1}, \\
    Xin Ding\textsuperscript{\rm 4}, Yongwei Wang\textsuperscript{\rm 1}\\
}
\affiliations{
    \textsuperscript{\rm 1} Zhejiang University \quad \quad \textsuperscript{\rm 2} Shanghai Maritime University \quad \textsuperscript{\rm 3} Zhejiang Gongshang University\\
    \textsuperscript{\rm 4} Nanjing University of Information Science and Technology \\
    \{lvkangtao, huangsen\_cao, yongwei.wang, zhimeng\}@zju.edu.cn\\
    kainantu03@126.com, yihuai1024@outlook.com, dingxin@nuist.edu.cn
}

\usepackage{bibentry}

\usepackage{color}

\begin{document}

\maketitle

\begin{abstract}
Large vision models have been found vulnerable to adversarial examples, emphasizing the need for enhancing their adversarial robustness. While adversarial training is an effective defense for deep convolutional models, it often faces scalability issues with large vision models due to high computational costs. Recent approaches propose robust fine-tuning methods, such as adversarial tuning of low-rank adaptation (LoRA) in large vision models, but they still struggle to match the accuracy of full parameter adversarial fine-tuning. The integration of various defense mechanisms offers a promising approach to enhancing the robustness of large vision models, yet this paradigm remains underexplored. To address this, we propose hyper adversarial tuning (HyperAT), which leverages shared defensive knowledge among different methods to improve model robustness efficiently and effectively simultaneously. Specifically, adversarial tuning of each defense method is formulated as a learning task, and a hypernetwork generates LoRA specific to this defense. Then, a random sampling and tuning strategy is proposed to extract and facilitate the defensive knowledge transfer between different defenses. Finally, diverse LoRAs are merged to enhance the adversarial robustness. Experiments on various datasets and model architectures demonstrate that HyperAT significantly enhances the adversarial robustness of pretrained large vision models without excessive computational overhead, establishing a new state-of-the-art benchmark.
\end{abstract}

\section{Introduction}
Transformers~\citep{vaswani2017attention} have set new benchmarks in diverse fields ranging from natural language processing to computer vision ~\citep{dosovitskiy2020image}. Open-source communities like Hugging Face and GitHub have made training data more accessible. Following the scaling law \citep{kaplan2020scaling}, leveraging large models pretrained on extensive datasets, followed by fine-tuning on downstream tasks, has become a prevalent paradigm in vision. However, these pretrained models are often vulnerable to adversarial attacks, some carefully crafted perturbations that can remarkably mislead models, thus raising significant security concerns in safety-critical applications \citep{wei2024physical}.

\begin{figure}[t]
    \includegraphics[width=0.45 \textwidth]{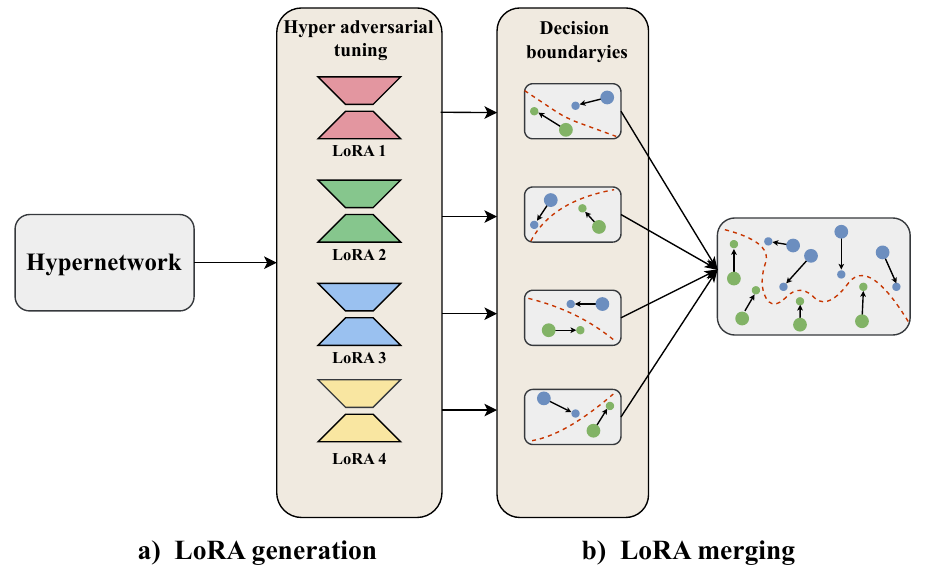}
    \centering
    \caption{The HyperAT framework involves two stages: a) generating weights for a mixture of defensive LoRAs, and b) merging weights to capture more sophisticated decision boundaries.}
    \label{fig:HyperAT FrameWork}
\end{figure}

In response, numerous typical defense methods have been proposed, including adversarial training~\citep{madry2018towards,zhang2019theoretically,cui2023decoupledkullbackleiblerdivergenceloss}, defensive distillation~\citep{cui2021learnable}, adversarial detection~\citep{roth2019odds}, and ensemble methods~\citep{wang2023exploring,croce2023seasoning}. Among these methods, adversarial training is widely recognized as a prominent approach, where adversarial examples are incorporated into the training process to increase the model's resilience to attacks. However, generating these adversarial samples requires multiple forward and backward propagations, which poses significant \textbf{\textit{efficiency}} challenges during adversarial training.

Efficient adversarial training approaches ~\citep{shafahi2019adversarial, wong2020fast} help alleviate computational costs. However, they often require full parameter fine-tuning, which still incurs substantial computational overhead for large models. 
Recently, Yuan et al introduce the FullLoRA-AT~\citep{yuan2024fulllora} framework by incorporating learnable LayerNorm LoRA \citep{hu2021lora} modules into ViT-based pretrained models. While this novel method enables rapid enhancement of adversarial robustness in a lightweight manner, its defensive effectiveness is inferior to its full-parameter adversarial fine-tuning counterpart.

Besides, existing adversarial training approaches often struggle with \textbf{\textit{effectiveness}} challenges, i.e., adversarially trained models may not be generalizable to unseen attacks~\citep{liu2022mutual, croce2023seasoning}. To address this issue, model soup~\citep{wortsman2022model} is introduced to improve robustness generalization by merging models trained on different types of attacks~\citep{croce2023seasoning}. Despite its defensive effectiveness, it requires separate training for each attack type, leading to significant overall training costs, particularly for large vision models.

To simultaneously address the \textbf{\textit{efficiency}} and \textbf{\textit{effectiveness}} challenges in conventional adversarial training for large vision models, we propose HyperAT, a novel robust tuning framework, by introducing a shared Hypernetwork~\citep{ha2016hypernetworks} and a mixture of defensive LoRAs into adversarial tuning. As illustrated in Fig.~\ref{fig:HyperAT FrameWork}, our method involves generating and merging stages. Fundamentally from existing adversarial tuning paradigms, the generating phase does not involve any learnable parameters specific to LoRAs, rather, a lightweight Hypernetwork is designed to generate weights for different LoRAs by formulating each defense as a learning task. Namely, we generate method-specific and layer-specific LoRA module parameters based on the defense method and layer ID embeddings. Thus, this design is very efficient in reducing computational costs and training time, particularly with many LoRAs. 

Besides, these defensive LoRAs, corresponding to diverse decision boundaries, are then merged into a single one. This process is beneficial to capture a smoother decision boundary, making our defense more generalizable than existing defenses. Moreover, inspired by Adamerging~\citep{yang2023adamerging}, we employ a simple yet effective approach to adaptively combine these diverse LoRA models, ultimately obtaining a more effective defense model named HyperAT+.

Extensive experiments using ViT-based large vision models on the CIFAR-10, CIFAR-100 and Imagenette datasets validate the efficiency and effectiveness of our method. Remarkably, HyperAT demonstrates superior robust accuracy compared to state-of-the-art parameter-efficient fine-tuning (PEFT) methods for adversarial robustness. It even surpasses the performance of fully fine-tuning the entire model while introducing significantly fewer learnable parameters.

In summary, our major contributions can be summarized as follows:
\begin{itemize}
\item We introduce HyperAT, a novel adversarial tuning framework for pretrained large vision models via Hypernetwork for defensive LoRA generation and model merging. Our method is parameter efficient, and it can facilitate knowledge transfer between different adversarial training methods. This significantly reduces the number of parameters required during adversarial training while improving model robustness. 
\item Our method is readily compatible and extensible to other adversarial training methods. By incorporating more advanced and powerful training methods, the overall model performance can be further enhanced without introducing excessive computational overhead.
\item With extensive experiments on three benchmark datasets, we demonstrate the superiority of HyperAT over existing state-of-the-art adversarial defenses. Notably, HyperAT can even surpass the robustness achieved by fully fine-tuning the entire model while requiring substantially fewer trainable parameters.
\end{itemize}
\section{Related Work}
\noindent\textbf{Adversarial Robustness.}
Despite the remarkable generalizability of large vision models pretrained on extensive datasets, they remain susceptible to adversarial attacks. A common defense mechanism is adversarial training, where deep neural networks are trained on crafted adversarial examples. A substantial body of research~\citep{zhang2019theoretically,wang2019improving} has been proposed to enhance the adversarial robustness of models. Chen et al.~\citep{chen2020adversarial}  were the first to introduce the concept of fine-tuning pretrained models to boost final model robustness. Along with approaches like RiFT~\citep{zhu2023improving}, AutoLoRa~\citep{xu2024autolora}, FullLoRA-AT~\citep{yuan2024fulllora}, and ARD \& PRM proposed by Mo et al.~\citep{mo2022adversarial}, these methods leverage pretraining and fine-tuning to achieve robust generalization.
Specifically, both AutoLoRa and FullLoRA-AT integrate LoRA~\citep{hu2021lora} during adversarial training. FullLoRA-AT incorporates LNLoRA modules into pretrained ViT models to achieve parameter-efficient robustness, while AutoLoRa separately optimizes natural and adversarial objectives by introducing LoRA branches, which helps avoid the instability often associated with simultaneously optimizing both objectives. Aforementioned works have demonstrated the effectiveness of combining pretrained models with adversarial tuning for boosting model robustness and reducing computational costs. Nevertheless, they tend to show compromised performances to unseen attacks.
\begin{figure*}[t]
    \includegraphics[width=0.85\textwidth]{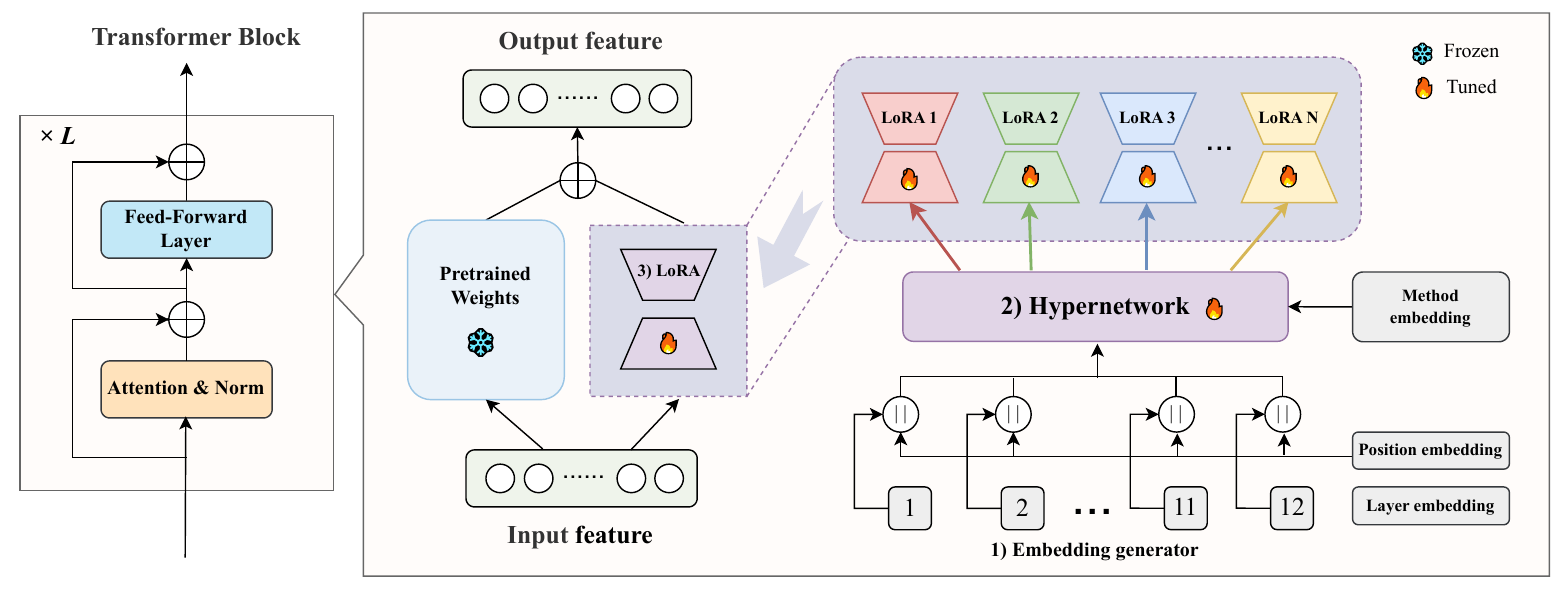}
    \centering
    \caption{Illustration of the proposed HyperAT architecture. The ViT model integrates HyperAT within both the attention and feed-forward blocks. HyperAT comprises three main components: 1) learned method embedding generator, 2) shared Hypernetwork, and 3) method-specific LoRA modules. We employ an embedding generator to produce method-specific embeddings for each adversarial training method. The shared hypernetwork then takes embeddings as input to generate the parameters for the method-specific LoRA module, which are used for adversarial fine-tuning with a small number of trainable parameters.}
    \label{fig:Model_Architecture}
\end{figure*}
\noindent\textbf{Parameter-efficient Finetuning and Hypernetworks.}
Several techniques have been proposed for parameter-efficient fine-tuning (PEFT) to reduce the number of trainable parameters, including Adapter~\citep{houlsby2019parameter}, Prefix-Tuning~\citep{li2021prefix}, Prompt Tuning~\citep{lester2021power} and LoRA. Among these, LoRA stands out by introducing a parallel low-rank adapter adjacent to the weights of a linear layer. By training only low-rank weight matrices, LoRA achieves performance comparable to full fine-tuning while significantly reducing the number of trainable parameters. Continuous efforts have been made to further enhance LoRA's efficiency, leading to various variants such as DyLoRA~\citep{valipour2022dylora}, AdaLoRA~\citep{zhang2023adalora}, QLoRA~\citep{dettmers2024qlora}, and LoRA-FA~\citep{zhang2023lora}.
However, when employing LoRA for multiple downstream tasks, each task requires its own specific LoRA training~\citep{huang2023lorahub}. As the number of tasks increases, the number of trainable parameters also inevitably grows. A promising solution to this challenge involves generating LoRA’s parameters using hypernetworks~\citep{ha2016hypernetworks}. Unlike traditional neural networks, hypernetworks do not directly process input data. Instead, they function as auxiliary networks that generate weights for a target network. Employing hypernetworks to produce adapter layers allows for knowledge sharing across multiple tasks, ensuring that the number of training parameters remains contained even as tasks proliferate, as demonstrated by HyperFormer~\citep{houlsby2019parameter}. While there have been recent studies combining LoRA and HyperNetworks in domains like image generation~\citep{ruiz2024hyperdreambooth} and physics-informed neural networks~\citep{majumdar2023hyperlora}, limited attention has been paid in the field of adversarial defense.

\noindent\textbf{Model Merging.}
Model merging aims to combine multiple pre-trained models into a more powerful or generalizable model, enabling it to perform multi-task learning. Recently, Wortsman et al. introduced Model Soups~\citep{wortsman2022model}, which interpolates the parameters of networks fine-tuned with different hyperparameter configurations from the same pre-trained model, thereby enhancing overall generalization. Similarly, Ilharco et al.~\citep{ilharco2022patching} fine-tuned a model across several image classification datasets and demonstrated that interpolating the original and fine-tuned parameters yields models that perform well across all tasks. However, inappropriate strategies for model merging can sometimes lead to performance degradation. To address this issue, various advanced methods have been developed in recent years to mitigate potential losses in performance. For instance, Fisher Merging~\citep{matena2022merging} and RegMean~\citep{jin2022dataless} use the Fisher information matrix and inner-product matrices to guide the merging process accordingly. Then, Task Arithmetic~\citep{ilharco2022editing} introduces the concept of task vectors, showing that merging these vectors effectively supports multitask learning. Building on this, PEM Composition~\citep{zhang2023composing} further integrates LoRA models into the task arithmetic framework, while Ties-Merging~\citep{yadav2306resolving} resolves task conflicts by resetting redundant parameters to address sign conflicts. To efficiently and effectively determine model merging coefficients, AdaMerging~\citep{yang2023adamerging} leverages an entropy minimization strategy on unlabeled test samples, iteratively refining the merging coefficients automatically.

\section{Method}
\subsection{Preliminary}
\textbf{Adversarial Training.}
Adversarial training typically involves the use of carefully crafted perturbations to enhance a model's robustness against adversarial attacks. The goal of generating adversarial examples \( x^{adv} \) is to find a perturbation \( \delta \) that maximizes the loss function \( \mathcal{L} \), while ensuring that the perturbation norm does not exceed \( \epsilon \) (i.e., \( \|\delta\|_\infty \leq \epsilon \)). This ensures that \( x^{adv} \) remains “close” to \( x \) but can still mislead the model into making an incorrect prediction (i.e., \( f(x+\delta;\theta) \neq y \)). Therefore, \( \delta \) can be estimated as follows:
\begin{equation}
    \delta = \arg\max_{\|\delta\|_p \leq \epsilon} \mathcal{L}(f(x+\delta;\theta), y)
\end{equation}
The $p$-norm can be 0, 1, 2, or \( \infty \). The loss function \( \mathcal{L} \) is typically the cross-entropy loss. Adversarial examples are defined as \( x^{adv} = x + \delta \). The objective is to train the model \( f(x;\theta) \) to minimize classification error against adversarial inputs. This is formulated as a mini-max problem ~\citep{huang2015learning}. Following ~\citep{shaham2018understanding}, the mathematical foundation of adversarial training can be described as follows:
\begin{equation}
    \min_\theta \mathbb{E}_{(x,y) \in \mathbb{D}} \left[ \max_{\|\delta\|_p \leq \epsilon} \mathcal{L}(f(x+\delta;\theta), y) \right]
\end{equation}

\noindent\textbf{Low-Rank Adaptation.}
Low-Rank Adaption (LoRA) ~\citep{hu2021lora} is a parameter-efficient fine-tuning technique that adapts a pretrained model to downstream tasks. LoRA achieves this by freezing most of the pretrained model’s weights $W_0 \in \mathbb{R}^{d \times k}$ and inserting trainable low-rank decomposition matrices to adjust the weights for adaptation. The forward computation of the adapted module is expressed as:

\begin{equation}
h = W_0x + \alpha \Delta Wx = W_0x + \alpha BAx
\end{equation}
where $B \in \mathbb{R}^{d \times r}$, $A \in \mathbb{R}^{r \times k}$, with $r \ll \min(d, k)$. Here, $r$ is a hyperparameter controlling the inner dimension of the matrices $A$ and $B$, balancing model adaptability and parameter efficiency. The hyperparameter $\alpha$ determines the influence of LoRA.

\noindent\textbf{HyperNetworks.}
Hypernetworks ~\citep{ha2016hypernetworks} are specialized neural networks that take an input, such as a simple vector or a latent representation, and output the weights $\theta$ for the primary network. Specifically, a hypernetwork denoted as $H(\cdot\ ;\phi)$ with independent parameters $\phi$, takes an embedding vector $\nu$ as input to generate the target parameters $\theta$. The process of generating the target parameters $\theta$ is as follows:
\begin{equation}
\theta = H(\nu;\phi)
\end{equation}

Let $M$ denote the total number of adversarial training methods and $m_\tau \in \mathbb{R}^t$ represent the method embedding corresponding to method-$\tau$. We compute individual method embeddings $\{m_\tau\}_{\tau=1}^M$ using a learned method projector network $h(\cdot\ ;I)$, which is a multi-layer perceptron consisting of two feed-forward layers and a ReLU non-linearity:
\begin{equation}
\nu_\tau = h(m_\tau;I)
\end{equation}
\subsection{HyperAT}
HyperAT is an innovative approach that combines hypernetworks with low-rank adaptation. As depicted in Fig. \ref{fig:Model_Architecture}, our method consists of three main components: 1) a learned method embedding generator, 2) method-specific LoRA modules, and 3) shared hypernetworks. We leverage an embedding generator to produce method embeddings $\nu_\tau$, specific to each adversarial training method. The shared hypernetwork then takes $\nu_\tau$ as input to generate the parameters for the method-specific LoRA module. Below, we will provide a detailed explanation of the method's intricacies.

Each layer of a Vision Transformer (ViT)~\citep{dosovitskiy2020image} model typically contains an attention block and a feed-forward block. In this paper, we primarily apply the method-specific LoRA to the weights in the \textit{Key}, \textit{Query}, and \textit{Value} parameters of the Attention block, as well as the two fully connected layers of the feed-forward block, which together comprise the majority of the parameters in ViT. As illustrated in Fig. \ref{fig:Model_Architecture}, our method-specific LoRA module can be seamlessly integrated into these blocks, emulating the block weights without altering the original model's structure or parameters.

Consistent with prior research ~\citep{mahabadi2021parameterefficientmultitaskfinetuningtransformers}, to enhance information sharing across each layer of the Transformer and improve parameter efficiency, we unify the method-specific LoRA across different layers and block types. In addition to the original task embeddings, we introduce layer ID embeddings $I = \{l_i\}_{i=1}^L$, which specify the block's layer, and block position embeddings $P = \{p_j\}_{j=1}^3$, indicating the specific network layer being substituted (e.g., the fully connected layers of the MLP block). This allows the shared hypernetwork to be reusable in generating the LoRA parameters for each method, position, and layer.
The method-specific LoRA modules $L_{\tau,I,p}$ include two low-rank decomposition matrices $A$ and $B$. The method-specific LoRA modules $L_{\tau,I,p}$ are defined as:
\begin{align}
    L_{\tau,I,p}
    &=\left(A_{\tau,I,p},B_{\tau,I,p}\right) \nonumber \\
    &=H_{\tau}\big(\nu_{\tau,I,p};\phi\big)
    =(W^{A},W^{B})\nu_{\tau,I,p},
\end{align}
where $\nu_{\tau,I,p} = h(m_\tau, I_i, p_j; I)$.

Here, $W^A$ and $W^B$ denote the weight matrices of the hypernetwork, which generate $A_{\tau,I,p}$ and $B_{\tau,I,p}$ for the $l$-th layer at the $p$-th position of the transformer block.

We enable the model to learn multiple adversarial training methods simultaneously. The hypernetwork acts as a general robustness information capturer across different methods, facilitating the transfer of generalizable knowledge. Unlike traditional continual learning, our approach does not suffer from catastrophic forgetting when learning multiple training methods concurrently and significantly reduces the computational cost typically required.

Specifically, during training, we randomly select one of the $M$ ($M \geq 2$) different adversarial training methods for iterative training, ultimately obtaining $M$ several specialists hypernetworks $H_1\left(\cdot \ ;\phi_1\right)$, $H_2\left(\cdot\ ;\phi_2\right)$, $\dots$, $H_M\left(\cdot\ ;\phi_M\right)$. Each LoRA is trained using a specific adversarial training method (e.g., PGD-AT ~\citep{madry2018towards}, TRADES ~\citep{zhang2019theoretically}), learning from the adversarial samples generated by that method and optimizing the parameters using the corresponding loss function.

During training, we only train the hypernetwork $H(\nu;\phi)$ and the method embedding generator $h(m_\tau;I)$, while keeping most of the pretrained model parameters $\theta$ fixed. In our work, we represent $m_\tau$ using the individual method name for clarity. For example, $x_{\text{pgd}}^{\text{adv}}$ denotes the adversarial samples generated while training the PGD-based specialist Hypernetwork $H_{\text{pgd}}\left(\cdot\ ;\phi_{\text{pgd}}\right)$. The loss function used during the training of $H_{\text{pgd}}\left(\cdot\ ;\phi_{\text{pgd}}\right)$ is $\mathcal{L}_{\text{pgd}}$ for PGD adversarial training.

After training, we obtain $M$ ($M \geq 2$) specific hypernetworks, each specialist network not only thoroughly trained on a specific method but also capable of extracting knowledge from other HyperATs through the shared hypernetwork. This allows the model to achieve better performance compared to training with a single method alone.

To further enhance the adversarial robustness, inspired by model soup ~\citep{croce2023seasoning}, we evenly merge the parameters generated by each specialist hypernetwork, thereby making the model even more robust to unseen attacks. The pseudocode for HyperLoRA-AT is summarized in Appendix A.

\subsection{HyperAT+}
Although model merging has been shown to effectively enhance model performance in numerous experiments, some research indicates its limitations. Adamergeing~\citep{yang2023adamerging} proposes an unsupervised adaptive model merging method that utilizes an entropy minimization strategy on unlabeled test samples as a surrogate optimization objective function to update the merging coefficient. Inspired by this, we use a few train samples rather than test samples to generate vanilla PGD-10 attack samples as inputs. Additionally, Adamergeing uses the entropy minimization strategy as an optimization objective in adversarial training, however, this may lead to degradation of robustness. To mitigate this issue, we adopt the following optimization form for the merging coefficients both Method-wise and Layer-wise, aiming to balance natural classification accuracy and robustness:
\begin{align}
    \min_{\{\lambda_1^l, \lambda_2^l, \dots, \lambda_m^l\}} \sum_{m=1}^{M} \sum_{x \in D} &\left( \mathcal{L}_{\mathrm{CE}}\left(f\left(x\right), y\right)\right. \nonumber\\
    &\left.+\lambda \cdot \mathcal{D}_{\mathrm{KL}}\left(f\left(x\right) \parallel f\left(x^{\mathrm{adv}}\right)\right) \right)
\end{align}
where $\lambda$ is a hyperparameter used to balance the importance of natural and robust errors. Throughout the entire process, all model parameters are frozen, and only the merging coefficients $\{\lambda_m^l\}_{m=1,l=1}^{M,L}$ are updated. This introduces minimal inference delay while significantly enhancing the overall robustness of the model. The algorithm refers to Appendix A.

\section{Experiments}
In this section, we introduce the specific details of our experiments, including the datasets, models, baselines, and evaluation metrics. Following this, we will present the main results, comparing the performance of HyperAT with various adversarial training methods used during full fine-tuning across different datasets. Additionally, we will compare our approach with existing state-of-the-art adversarial tuning methods on pretrained large vision models. Following this, we will analyze the parameter efficiency of HyperAT. Finally, we conduct several ablation studies to explore the impact of the major components.
\subsection{Experimental Setup}
\noindent\textbf{Datasets and Models.} 
We conduct experiments on the CIFAR-10~\citep{krizhevsky2009learning}, CIFAR-100~\citep{krizhevsky2009learning}, and Imagenette~\citep{Jere2019imagenette} datasets, which are popular for adversarial training~\citep{croce2020robustbench}. The CIFAR-10 with 10 classes and CIFAR-100 with 100 classes are subsets of the Tiny Images dataset, with training and test sets containing 50,000 and 10,000 images respectively. Imagenette is a subset of 10 easily classified classes from the ImageNet dataset, consisting of 9,469 training images and 3,925 test images, each of size 224×244. Additionally, we primarily conduct experiments using ViT-B~\citep{dosovitskiy2020image}, ViT-L~\citep{dosovitskiy2020image}, and DeiT~\citep{pmlr-v139-touvron21a}.

\noindent\textbf{Baseline Methods.}
We use the performance of vanilla AT, TRADES~\citep{zhang2019theoretically}, MART~\citep{wang2019improving}, and the recently proposed DKL~\citep{cui2023decoupledkullbackleiblerdivergenceloss}, which update the entire set of model parameters, as our baselines. Additionally, we compare our method with several state-of-the-art parameter-efficient fine-tuning (PEFT) methods for adversarial robustness, namely, LoRA~\citep{hu2021lora}, Aurora~\citep{wang2023parameter}, FullLoRA-AT~\citep{yuan2024fulllora}, and AutoLoRA~\citep{xu2024autolora}.

\noindent\textbf{Evaluation Metrics.}
To compare the performance of different methods, we evaluate the model's adversarial robustness using PGD-20, CW-20~\citep{carlini2017towards}, and AutoAttack (AA)~\citep{croce2020reliable}, with an adversarial budget of 8/255. Additionally, we calculate the standard test accuracy and the average accuracy across the aforementioned evaluation metrics to assess the trade-off between clean accuracy and adversarial robustness for these methods.

\noindent\textbf{Experiment Details.}
For a fair comparison, all our experiments were conducted over 40 epochs. We utilized the SGD optimizer, with the weight decay fixed at $1e^{-4}$. The learning rate, initially set to 0.1, was scaled down by a factor of 0.1 after the 28th and 36th epochs. During training, we employed standard adversarial training with a PGD-10 attack, using an adversarial budget of 8/255 and a step size of 2/255 to generate adversarial perturbations. All adversaries type = $l_\infty$. For LoRA-based methods, the rank r was set to 16 by default. For CIFAR10 and CIFAR100 datasets, the batch size for all experiments is set to 256, and for Imagenette datasets, the batch size is 64. All experiments were conducted on a server with two NVIDIA GeForce RTX 4090 GPUs.

\subsection{Results and Analysis}
\textbf{Main Results.}
In Tables \ref{tab:Comparison with existing adversarial training methods with fully fine-tuning across different datasets using ViT-B} and \ref{tab:Comparison with state-of-the-art PEFT methods for enhancing adversarial robustness on pretrained models}, we demonstrate that our method, HyperAT, significantly outperforms existing adversarial training methods with fully fine-tuned model parameters and state-of-the-art PEFT methods for enhancing adversarial robustness on pretrained models. 

HyperAT integrates four adversarial training strategies—vanilla AT, MART, TRADES, and DKL—during the training process. The results clearly indicate that HyperAT consistently achieves higher robust test accuracy compared to employing a single adversarial defense strategy. Furthermore, when benchmarked against methods that require updating the entire set of model parameters to improve robustness, HyperAT demonstrates superior robustness while significantly reducing the number of trainable parameters.

As illustrated in Table \ref{tab:Comparison with existing adversarial training methods with fully fine-tuning across different datasets using ViT-B}, Vanilla AT consistently demonstrates stable improvements in robustness across all datasets. However, when compared to Vanilla AT, HyperAT boosts overall robust accuracy by approximately 7–8\% under various attack evaluations. Particularly for the CIFAR-100 dataset, HyperAT significantly enhances model robustness without sacrificing standard test accuracy. This improvement is attributed to the fine-grained nature of CIFAR-100 classes, where the higher intra-class variations present a challenge for the model to learn robust features. By integrating multiple defense methods during training, HyperAT is able to generate more effective adversarial examples within each method, thereby leading to a stronger robust generalization of the model.
\begin{table}[t]
    \centering
    \caption{\textbf{Comparison with existing adversarial training methods with fully fine-tuning across different datasets using ViT-B.} “Clean Acc” refers to the standard test accuracy. “PGD-20”, “CW-20” and “AA” refer to the robust test accuracy evaluated by PGD-20, CW-20 and AutoAttack, respectively. “Average Acc” represents the average of all evaluation metrics. The best results are in bold and the second-best is underlined.}
    \resizebox{\columnwidth}{!}{
        \begin{tabular}{c|c|c|cccc|c}
            \toprule
            \multirow{2}[2]{*}{Dataset} & \multirow{2}[2]{*}{Method} & Trainable & Clean Acc & PGD-20 & CW-20 & AA & Average Acc \\
                &     & Pars (M) & (\%)  & (\%)  & (\%)    & (\%)  & (\%) \\
            \midrule
            \multirow{7}[5]{*}{CIFAR-10} & Standard Training& 85.15& 97.52&	0.00& 0.00&	0.00& 24.38\\
            \cmidrule{2-8}
            & Vanilla AT (\citeyear{madry2018towards})& 85.15&	\textbf{87.22}&	50.25& 49.51& 48.55& 58.88\\
            & MART (\citeyear{wang2019improving})& 85.15& 83.45& 51.59& \textbf{54.64}& 47.15& 59.21\\
            & TRADES (\citeyear{zhang2019theoretically})& 85.15&	85.70&	49.94& 50.44& 48.09& 58.54 \\
            & DKL (\citeyear{cui2023decoupledkullbackleiblerdivergenceloss})& 85.15& 85.11&	51.59& 51.38&	49.21& 59.32 \\
            \cmidrule{2-8}
            & \textbf{HyperAT(ours)}& 18.26& 85.54& \underline{53.93}&	51.81& \underline{50.29}& \underline{60.39} \\
            \cmidrule{2-8}
            & \textbf{HyperAT+ (ours)}& 18.26&	\underline{85.96}& \textbf{54.66}& \underline{51.97}& \textbf{50.47}& \textbf{60.77} \\
            \midrule
            \multirow{7}[5]{*}{CIFAR-100} & Standard Training& 85.15& 90.71& 0.00& 0.00& 0.00& 22.68\\
            \cmidrule{2-8}
            & Vanilla AT (\citeyear{madry2018towards})& 85.15&	\underline{65.43}&	27.65&	27.65&	25.90&	36.66\\
            & MART (\citeyear{wang2019improving})& 85.15&	56.34&	25.50&	25.05&	23.06&	32.49\\
            & TRADES (\citeyear{zhang2019theoretically})&85.15&	63.15&	26.70&	27.81&	23.96&	35.41\\
            & DKL (\citeyear{cui2023decoupledkullbackleiblerdivergenceloss})&85.15&	62.21&	27.87&	28.22&	26.22&	36.23\\
            \cmidrule{2-8}
            & \textbf{HyperAT(ours)}& 18.26&	\textbf{66.04}&	\underline{31.32}&	\underline{29.01}&	\underline{27.22}&	\underline{38.54}\\
            \cmidrule{2-8}
            & \textbf{HyperAT+ (ours)}& 18.26&	65.33&	\textbf{32.18}&	\textbf{29.04}&	\textbf{27.58}&	\textbf{38.53}\\
            \midrule
            \multirow{7}[5]{*}{Imagenette} & Standard Training& 85.15&	99.14&	0.00&	0.00&	0.00&	24.78\\
            \cmidrule{2-8}
            & Vanilla AT (\citeyear{madry2018towards})&85.81&	87.89&	66.41&	64.06&	60.42&	69.70\\
            & MART (\citeyear{wang2019improving})&85.81&	88.67&	67.97&	\textbf{68.75}&	61.72&	71.78\\
            & TRADES (\citeyear{zhang2019theoretically})& 85.81&	\underline{89.65}&	63.70&	62.86&	61.81&	69.51\\
            & DKL (\citeyear{cui2023decoupledkullbackleiblerdivergenceloss})&85.81&	87.50&	63.39&	62.95&	62.50&	69.09\\
            \cmidrule{2-8}
            & \textbf{HyperAT(ours)}&18.92& 88.23&	\underline{69.53}&	66.41&	\underline{65.10}&	\underline{72.32}\\
            \cmidrule{2-8}
            & \textbf{HyperAT+ (ours)}&18.92&	\textbf{90.62}&	\textbf{71.09}&	\underline{67.58}&	\textbf{65.89}&	\textbf{73.80}\\
            \bottomrule
        \end{tabular}%
    }
    \label{tab:Comparison with existing adversarial training methods with fully fine-tuning across different datasets using ViT-B}%
\end{table}%
Moreover, while methods like LoRA, Aurora, and FullLoRA-AT employ a small number of additional parameters to achieve a robust model, they still fall short of matching the accuracy achieved through full parameter adversarial fine-tuning. The AutoLoRA method, which optimizes natural objectives via the LoRA branch and adversarial objectives through the feature extractor, manages to avoid the instability often associated with simultaneously optimizing both objectives using the full feature extractor. However, it still requires fine-tuning the feature extractor alongside updating the LoRA weights, leading to unsatisfactory training times.
\begin{table}[t]
    \centering
    \caption{\textbf{Comparison with state-of-the-art PEFT methods for enhancing adversarial robustness on pretrained models.} Benchmark methods are based on ViT-B and fine-tuned on CIFAR-10 dataset using Vanilla AT. “$\Delta$” represents the cumulative difference in performance across all evaluation metrics compared to fully fine-tuning with Vanilla AT.}
    \resizebox{\columnwidth}{!}{
        \begin{tabular}{c|c|cccc|c}
            \toprule
             \multirow{2}[2]{*}{Method} & Trainable & Clean Acc & PGD-20 & CW-20 & AA & $\Delta$ \\
                & Pars (M) & (\%)  & (\%)  & (\%) & (\%) & (\%) \\
            \midrule
            Vanilla AT (\citeyear{madry2018towards})&	85.15&	87.22&	50.25&	49.51&	48.55&	- \\
            \cmidrule{1-7}
            LoRA (\citeyear{hu2021lora})&	9.36&	87.87&	48.63&	48.16&	47.25&	-3.62\\
            Aurora (\citeyear{wang2023parameter})& 7.56&	87.21& 50.70& 49.42& 47.10& -1.10\\
            FullLoRA-AT(\citeyear{yuan2024fulllora})& 9.40&	87.62&	50.96&	49.84&	47.14&	+0.03\\
            AutoLoRA (\citeyear{xu2024autolora})& 87.51& 80.70& 52.46& 47.65& 46.44& -8.28\\		
            \cmidrule{1-7}
            \textbf{HyperAT (ours)}&	18.26&	85.54&	53.93&	51.81&	50.29&	+6.04\\
            \textbf{HyperAT+ (ours)}& 18.26&	85.96&	54.66&	51.97&	50.47&	+7.53\\
            \bottomrule
        \end{tabular}%
    }
    \label{tab:Comparison with state-of-the-art PEFT methods for enhancing adversarial robustness on pretrained models}%
\end{table}%

\subsection{Parameter Efficiency Analysis}
In this section, we compare the computational, storage efficiency and robustness between LoRA and HyperAT.

\noindent\textbf{Computational Efficiency.}
When aiming to robustly fine-tune a pretrained model using a specific adversarial training method, we typically need to train the model for T epochs to obtain a robust model along with its corresponding LoRA module. In scenarios where different tasks or datasets require distinct training methods, each task necessitates an individual robust fine-tuning of the pretrained model. Consequently, the training time scales linearly with the number of methods; for M training methods, a total of M·T epochs is required. In contrast, HyperAT enables the simultaneous learning of M methods within the same T epochs, significantly reducing the overall training time required for multiple adversarial training tasks while still producing a highly robust model. Moreover, during training, positive knowledge transfer occurs between the learning tasks, allowing the LoRA modules generated by each method to surpass the performance of models fully fine-tuned using individual adversarial training methods. The detailed results are provided in Appendix B.

\noindent\textbf{Storage Efficiency.}
Training large vision models requires significant computational and storage resources, making the cost of robust fine-tuning these models prohibitively high. LoRA offers a more efficient alternative to full fine-tuning; however, as the number of fine-tuning tasks for specific downstream applications increases, the storage and deployment resources required for each individual LoRA module also grow. HyperAT effectively addresses this issue by utilizing a shared hypernetwork to generate LoRA modules tailored to specific tasks. Although HyperAT initially requires more parameters than a single LoRA module, as the number of required LoRA modules increases, the additional parameters needed for the shared hypernetwork become negligible. This allows HyperAT to generate hundreds or even thousands of LoRA modules using a single network without introducing significant additional storage costs.

\noindent\textbf{Robustness Generalization.}
In terms of model robustness, a single LoRA module trained using a specific adversarial method may achieve a local optimum tailored to a particular type of attack but remains vulnerable to other, unknown attack types. Our method leverages the strengths of multiple defense strategies, merging the resulting local optima, which effectively broadens the flat wide minima—referring to the loss surface around the minima. The flatter and wider this loss surface, the better the model's generalization performance. 

\subsection{Ablation Studies}
In this subsection, we conducted ablation studies on the various models, rank r, the number of Methods combined during robust training, and the number of HyperAT+ iterations.

\begin{table}[t]
    \centering
    \caption{\textbf{Comparison with existing adversarial training methods using different ViT-based models on the CIFAR-10 dataset.}}
    \resizebox{\columnwidth}{!}{
        \begin{tabular}{c|c|c|cccc|c}
            \toprule
            \multirow{2}[2]{*}{Model} & \multirow{2}[2]{*}{Method} & Trainable & Clean Acc & PGD-20 & CW-20 & AA & Average Acc \\
                &   & Pars (M) & (\%) & (\%)  & (\%) & (\%) & (\%) \\
            \midrule
            \multirow{7}[5]{*}{VIT-B} & Standard Training& 85.15&	97.52&	0.00&	0.00&	0.00&	24.38\\
            \cmidrule{2-8}
            & Vanilla AT (\citeyear{madry2018towards})& 85.15&	\textbf{87.22}&	50.25&	49.51&	48.55&	58.88\\
            & MART (\citeyear{wang2019improving})& 85.15&	83.45&	51.59&	\textbf{54.64}&	47.15&	59.21\\
            & TRADES (\citeyear{zhang2019theoretically})&85.15&	85.70&	49.94&	50.44&	48.09&	58.54\\
            & DKL (\citeyear{cui2023decoupledkullbackleiblerdivergenceloss})& 85.15&	85.11&	51.59&	51.38&	49.21&	59.32\\
            \cmidrule{2-8}
            & \textbf{HyperAT (ours)}&18.26&	85.54&	\underline{53.93}&	51.81&	\underline{50.29}&	\underline{60.39}\\
            \cmidrule{2-8}
            & \textbf{HyperAT+ (ours)}&18.26&	\underline{85.96}&	\textbf{54.66}&	\underline{51.97}&	\textbf{50.47}&	\textbf{60.77}\\
            \midrule
            \multirow{7}[5]{*}{VIT-L} & Standard Training&302.43& 97.50& 0.00& 0.00& 0.00& 24.38\\
            \cmidrule{2-8}
            & Vanilla AT (\citeyear{madry2018towards})&302.43& \textbf{89.99}& 51.31& 51.53& 49.59& 60.61\\
            & MART (\citeyear{wang2019improving})&302.43& 84.75& 50.02& \textbf{53.87}& 46.67& 58.83\\
            & TRADES (\citeyear{zhang2019theoretically})&302.43& 85.86& 50.33& 51.00& 48.53& 58.93\\
            & DKL (\citeyear{cui2023decoupledkullbackleiblerdivergenceloss})&302.43& 84.02& \textbf{53.99}& 52.72& \textbf{51.13}& 60.47\\
            \cmidrule{2-8}
            & \textbf{HyperAT(ours)}&40.13& 88.83& 53.73& 52.88& 50.87& \underline{61.58}\\
            \cmidrule{2-8}
            & \textbf{HyperAT+ (ours)}&40.13& \underline{88.99}& \underline{53.86}& \underline{52.91}& \underline{51.04}& \textbf{61.70}\\
            \midrule
            \multirow{7}[5]{*}{DeiT-B} & Standard Training&85.17& 97.84& 0.00& 0.00& 0.00& 0.00\\
            \cmidrule{2-8}
            & Vanilla AT (\citeyear{madry2018towards})&85.17& \textbf{88.11}& 50.87& 49.82& 48.08& 59.00\\
            & MART (\citeyear{wang2019improving})&85.17& 82.46& 50.97& 48.52& 46.61& 57.12\\
            & TRADES (\citeyear{zhang2019theoretically})&85.17& \underline{85.61}& 50.51& 50.39& 48.51& 58.76\\
            & DKL (\citeyear{cui2023decoupledkullbackleiblerdivergenceloss})&85.17& 82.94& 53.39& 50.60& 49.68& 59.15\\
            \cmidrule{2-8}
            & \textbf{HyperAT (ours)}& 18.28& 84.44& \underline{53.71}& \underline{51.52}& \underline{50.08}& \underline{59.94}\\
            \cmidrule{2-8}
            & \textbf{HyperAT+ (ours)}& 18.28& 84.30& \textbf{54.00}& \textbf{51.63}& \textbf{50.20}& \textbf{60.03}\\
            \bottomrule
        \end{tabular}%
    }
    \label{tab:Comparison with existing adversarial training methods using different ViT-based models on the CIFAR-10 dataset}%
\end{table}%

\noindent\textbf{HyperAT in different ViT-based model.}
We conducted several experiments on different ViT-based models using the CIFAR-10 dataset to validate the effectiveness of our method. The results are presented in Table \ref{tab:Comparison with existing adversarial training methods using different ViT-based models on the CIFAR-10 dataset}. Consistent with our previous findings, our approach demonstrates significant superiority across various models. Notably, for larger vision models such as ViT-L, the proportion of additional trainable parameters introduced by HyperAT is even smaller compared to that in ViT-B, which decreases from 18.98\% to 12.65\%. This indicates that as the model scale increases, the parameter efficiency advantage of HyperAT becomes even more pronounced.

\begin{table}[t]
    \centering
    \caption{\textbf{Effects of different rank \textit{r} on the performance of HyperAT.} Experiments were conducted using ViT-B on the CIFAR-10 dataset. “Param. Ratio” denotes the ratio of trainable parameters to the original model parameters.}
    \resizebox{\columnwidth}{!}{
        \begin{tabular}{c|c|c|cccc|c}
            \toprule
             \multirow{2}[2]{*}{Method} & Rank &Param.&Clean Acc& PGD-20 & CW-20 & AA & $\Delta$ \\
               & ($r$) & Ratio & (\%)  & (\%)  & (\%)    & (\%)  & (\%) \\
            \midrule
            Vanilla AT &- &	100\%&	87.22&	50.25&	49.51&	48.55&	-\\
            \cmidrule{1-8}
            FullLoRA-AT& 16& 10.76\%& 87.62& 50.96&	49.84&	47.14&	+0.03\\
            \cmidrule{1-8}
            \multirow{3}[4]{*}{\textbf{HyperAT(ours)}} &8& 14.07\%& 86.86&	52.82&	51.42&	49.58&	+5.15\\
            &16& 18.98\%& 85.54&	53.93&	51.81&	50.29&	+6.06\\
            &32& 27.30\%& 87.83&	53.55&	52.49&	50.51&	+8.85\\
            &64& 39.69\%& 87.38&	53.50&	52.43&	50.45&	+8.23\\
            \bottomrule
        \end{tabular}%
    }
    \label{tab:Effect of different rank r on the performance of HyperAT}%
\end{table}%

\noindent\textbf{Effects of the Rank of LoRA.}
Table \ref{tab:Effect of different rank r on the performance of HyperAT} shows that both test accuracy and robustness generally improve as the rank $r$ increases. This is because, as $r$ grows, the hypernetwork can generate LoRA modules with a larger parameter space, effectively simulating the original network layers and achieving results comparable to full fine-tuning. However, this does not imply that a larger $r$ is always better. In our experiments, we observed that while increasing $r$ from 8 to 64, the performance gain becomes marginal when $r=32$, indicating that $r=32$ is sufficient to capture the critical robust features. Considering the trade-off between the number of parameters used for adversarial fine-tuning and the robustness of the model, we selected $r=16$ as the default hyperparameter.

\begin{table}[thbp]
    \centering
    \caption{\textbf{Effects of the different number of methods combined in HyperAT during training.} Experiments were conducted using ViT-B on the CIFAR-10 dataset.}
    \resizebox{\columnwidth}{!}{
        \begin{tabular}{c|cccc|c}
            \toprule
             The numbers of   & Clean Acc & PGD-20 & CW-20 & AA & $\Delta$ \\
             methods combined& (\%)  & (\%)  & (\%)    & (\%)  & (\%) \\
            \midrule
            Vanilla AT&87.22&	50.25&	49.51&	48.55&	-\\
            \cmidrule{1-6}
            FullLoRA-AT& 87.62& 50.96&	49.84&	47.14&	+0.03\\
            \cmidrule{1-6}
            2 methods&	85.68&	52.22&	50.61&	48.21&	+1.19\\
            3 methods&	86.36&	52.32&	50.83&	49.67&	+3.65\\
            4 methods&	85.54&	53.93&	51.81&	50.29&	+6.06\\
            5 methods&	85.55&	54.18&	52.03&	50.59&	+6.82\\
            \bottomrule
        \end{tabular}%
    }
    \label{tab:Effect of the number of Methods}%
\end{table}%

\noindent\textbf{Effects of the Number of Defense Methods.}
HyperAT supports the extension to multiple adversarial training methods, enhancing the effectiveness of existing approaches and improving overall model robustness as new, effective adversarial training techniques are incorporated. To validate the impact of the number of methods used in HyperAT on overall model robustness, we conducted experiments with varying numbers of methods, as shown in Table \ref{tab:Effect of the number of Methods}. Specifically, when the number of methods is set to two, we combine Vanilla AT and MART for training. When the number increased to three, we added TRADES to the existing combination of Vanilla AT and MART. With four methods, we further incorporate DKL, which is the default configuration of our proposed approach.
To further explore the extensibility of our method, we introduced the SCORE method in addition to the previous four. SCORE (Self-Consistent Robust Error)~\citep{pmlr-v162-pang22a}is an adversarial training approach that redefines robust error to better balance robustness and accuracy. It replaces local invariance with local equivariance and utilizes distance metrics instead of KL divergence, aligning model predictions more closely with the true data distribution and addressing the robustness-accuracy trade-off more effectively. 

The inclusion of SCORE further improved the overall robustness of HyperAT, demonstrating its strong compatibility with additional methods. However, as the number of methods increases, the incremental benefit to robustness diminishes, showing diminishing marginal returns. Considering computational efficiency and algorithmic complexity, we chose four methods as the default training setup for our work. Additionally, we discovered that the specific combination of adversarial training methods during training also impacts the enhancement of overall robustness. Detailed results are available in Appendix C. We found that combining the most effective methods can yield a more robust model, and the overall performance is somewhat influenced by methods that perform less well on specific datasets. Nonetheless, HyperAT consistently outperforms any single method in training. This flexibility in the method combination of training offers greater potential for enhancing model robustness.

\noindent\textbf{Effect of the Number of adjustment Iterations.}
For our method HyperAT+, as shown in Appendix D, we observe that model performance tends to decline as the number of HyperAT+ adjustment iterations increases. This decline is primarily due to the impact of PGD attacks, where the model's original decision boundaries overfit these perturbed samples, thereby destabilizing the previously robust decision boundaries. To balance robustness performance with training efficiency, we have selected 7 iterations as the default setting for adjustment.

\section{Conclusion}
In this work, we have introduced a novel adversarial tuning framework for large pretrained vision models, entitled HyperAT. By utilizing a lightweight hypernetwork to generate LoRA weights specific to diverse defense methods, our method facilitates defensive knowledge transfer between diverse adversarial training methods. HyperAT significantly reduces the number of parameters required during adversarial training while substantially boosting the model's robustness. In addition, we proposed a flexible merging strategy HyperAT+ to leverage multiple LoRAs generated by HyperAT. The merging procedure helps capture smoother decision boundaries, thus making our defense more generalizable to unseen attacks. Comprehensive empirical results demonstrate that our approach can significantly enhance the adversarial robustness of pretrained large vision models, meanwhile maintaining high computational efficiency.

\appendix 
\section{Appendix A}
\begin{algorithm}[thbp]
\caption{HyperAT algorithm}
\label{alg:HyperAT algorithm}
\textbf{Input}: Training samples $(\mathcal{X},\mathcal{Y}) \in \mathbb{D}$, model \( f(\cdot ; \theta) \), method embeddings $\{m_\tau\}_{\tau=1}^M$,  embedding generator $h(\cdot\ ;I)$ and Hypernetwork $H(\cdot\ ;\phi)$ \\
\textbf{Parameter}: $\theta,I,\phi$, learning rate $\alpha$\\
\textbf{Output}: specialist LoRA weight
\begin{algorithmic}[1]
\STATE freezing most of pretrained weight $\theta$
\FOR{epoch = 1 to $N$}
    \FOR{minibatch $(x, y) \subseteq \mathcal{X} \times \mathcal{Y}$}
        \STATE Randomly select one adversarial training method $\tau$\\
        \STATE Generate adversarial examples of $\tau:x^{\text{adv}}_{\tau} = \textit{attack}(x, y, f(\cdot ; \theta),H(h(\cdot\ ;I) ;\phi))$
        \STATE Compute loss: $\mathcal{L}_{\tau}$
        \STATE Update $\theta,I,\phi$ with gradient descent: $(\theta,I,\phi) \gets (\theta,I,\phi) - \alpha \nabla_{(\theta,I,\phi)} \mathcal{L}_{\tau}$
    \ENDFOR
\ENDFOR\\
\IF {mode = merge}
\STATE Merging every specialist LoRA weight
\ENDIF
\STATE \textbf{return} 
\end{algorithmic}
\end{algorithm}

\begin{algorithm}[h!]
\caption{HyperAT+ algorithm}
\label{alg:HyperAT+ algorithm}
\textbf{Input}: Training samples $(\mathcal{X},\mathcal{Y}) \in \mathbb{D}$, model \( f(\cdot ; \theta) \), method embeddings $\{m_\tau\}_{\tau=1}^M$,  embedding generator $h(\cdot\ ;I)$ and Hypernetwork $H(\cdot\ ;\phi)$ \\
\textbf{Parameter}: $\theta,I,\phi$, learning rate $\alpha$, Layer-wise and Method-wise $\lambda$\\
\textbf{Output}:$\lambda$
\begin{algorithmic}[1] 
\STATE Freezing $\theta,I,\phi$
\FOR{epoch = 1 to $N$}
    \FOR{minibatch $(x, y) \subseteq \mathcal{X} \times \mathcal{Y}$}
        \STATE Generate adversarial examples of PGD attack: $x^{\text{adv}} = \textit{attack}(x, y, f(\cdot ; \theta),H(h(\cdot\ ;I) ;\phi), \lambda)$
        \STATE Compute loss: $\mathcal{L}=\mathcal{L}_{\mathrm{CE}}\left(f\left(x\right),y\right)+\lambda \cdot \mathcal{D}_{\mathrm{KL}}(f\left(x\right)\parallel(x^{\mathrm{adv}}))$
        \STATE Update $\lambda$ with gradient descent: $\lambda \gets \lambda - \alpha \nabla_{\lambda} \mathcal{L}$
\IF {minibatch $>$ 1}
        \STATE break
        \ENDIF
    \ENDFOR
\ENDFOR\\
\STATE \textbf{return} 
\end{algorithmic}
\end{algorithm}\vspace{1.5in}

\section{Appendix B}
As shown in Table 6, we compare the performance of full fine-tuning, LoRA and method-specific LoRA modules generated by HyperAT. We find that each method-specific LoRA module surpasses the performance of models  using individual adversarial training method. Additionally, we performed robust fine-tuning on ViT-B using LoRA with each defense method on the CIFAR-10 dataset. We then merged the weight of independently trained LoRA modules and compared this ensemble, labeled as LoRA (ensemble) in the table, with our HyperAT method. The results demonstrate that HyperAT exhibits significant superiority, indicating that positive knowledge transfer occurs between the learning tasks while effectively broadening the flat wide minima during model merging.

\begin{table}[h!]
    \centering
    \caption{The performance of the method-specific LoRA modules generated by HyperAT.}
    \resizebox{\columnwidth}{!}{
        \begin{tabular}{c|c|cccc|c}
            \toprule
             \multirow{2}{*}{Method} & Trainable & Clean Acc & PGD-20 & CW-20 & AA & AVG \\
                & Pars (M) & (\%)  & (\%)  & (\%) & (\%) & (\%) \\
            \midrule
            Vanilla AT (Fully Finetune)&	85.15&	87.22&	50.25&	49.51&	48.55&	58.88 \\
            Vanilla AT (LoRA)&	9.36&	87.87&	48.63&	48.16&	47.25&	57.98 \\
            HyperAT ($\tau$ = Vanilla AT)&	18.26&	87.11&	52.17&	51.32&	49.15&	59.95\\
            \cmidrule{1-7}
            MART (Fully Finetune)&	85.15&	83.45&	51.59&	54.64&	47.15&	59.21 \\
            MART (LoRA)&	9.36&	82.44&	52.30&	51.22&	47.18&	58.29 \\
            HyperAT ($\tau$ = MART)& 18.26&	84.76&	53.42&	51.56&	49.49&	59.81\\
            \cmidrule{1-7}
            TRADES (Fully Finetune)&	85.15&	85.70&	49.94&	50.44&	48.09&	58.54 \\
            TRADES (LoRA)&	9.36&	86.37&	51.69&	50.81&	49.07&	59.49 \\
            HyperAT ($\tau$ = TRADES)& 18.26&	85.30&	53.30&	51.15&	49.85&	59.90\\
            \cmidrule{1-7}
            DKL (Fully Finetune)&	85.15&	85.11&	51.59&	51.38&	49.21&	59.32 \\
            DKL (LoRA)&	9.36&	83.45&	54.21&	51.81&	50.40&	60.04 \\
            HyperAT ($\tau$ = DKL)& 18.26&	83.0&	54.46&	51.40&	50.21&	59.78\\
            \cmidrule{1-7}
            LoRA (ensemble)&	-&	88.76&	50.81&	48.89&	46.70&	58.79\\
            \textbf{HyperAT (ours)}&	18.26&	85.54&	53.93&	51.81&	50.29&	60.39\\
            \textbf{HyperAT+ (ours)}& 18.26&	85.96&	54.66&	51.97&	50.47&	60.77\\
            \bottomrule
        \end{tabular}%
    }
    \label{tab:The performance of method-specific LoRA module generated by HyperAT}%
\end{table}

\section{Appendix C}
As illustrated in Table 7, we experimented with different combinations of defense methods during HyperAT training. For example, Vanilla AT + MART indicates that only Vanilla AT and MART were combined during the training process. The effectiveness of different combinations of adversarial training methods varies in terms of enhancing robustness. Overall, incorporating the most innovative defense methods during training tends to provide a greater boost to model robustness. Additionally, the flexibility in combining these methods offers further potential for enhancement.

\begin{table}[h!]
    \centering
    \caption{The performance of different method combinations that hyperAT used during training}
    \resizebox{\columnwidth}{!}{
        \begin{tabular}{c|cccc|c}
            \toprule
             \multirow{2}[2]{*}{combination}& Clean Acc& PGD-20& CW-20& AA & AVG \\
                        & (\%)  & (\%)  & (\%)   & (\%)  & (\%) \\
            \midrule
            Vanilla AT + MART&85.68&	52.22&	50.61&	48.21&	59.18\\
            Vanilla AT + TRADES&86.56&	52.93&	51.42&	49.60&	60.13\\
            Vanilla AT + DKL&  83.10&   54.13&	51.81&	50.38&  59.86\\
            \cmidrule{1-6}
            Vanilla AT + MART + TRADES& 86.36& 52.32& 50.83& 49.67&	59.80\\
            Vanilla AT + MART + DKL& 84.27& 54.36&	51.92&	50.04&	60.15\\
            Vanilla AT + TRADES + DKL& 86.20& 53.83&	51.74&	50.15&60.48\\
            MART+ TRADES + DKL& 54.81& 54.20&	51.78&	50.31&60.28\\
            \cmidrule{1-6}
            Vanilla AT + MART +TRADES +DKL &	85.54&	53.93&	51.81&	50.29&	60.39\\
            \bottomrule
        \end{tabular}%
    }
    \label{tab:deifferent combination of Methods}%
\end{table}%

\clearpage
\section{Appendix D}
As shown in Figure 3, it can be observed that as the number of HyperAT+ adjustment iterations increases, the model's performance tends to decline. This decline is primarily due to the impact of PGD attacks and the small value of the parameter $\lambda$, which causes the model's original decision boundaries to overfit these perturbed samples, thereby destabilizing the previously robust decision boundaries.

\begin{figure}[h!]
    \includegraphics[width=0.45\textwidth]{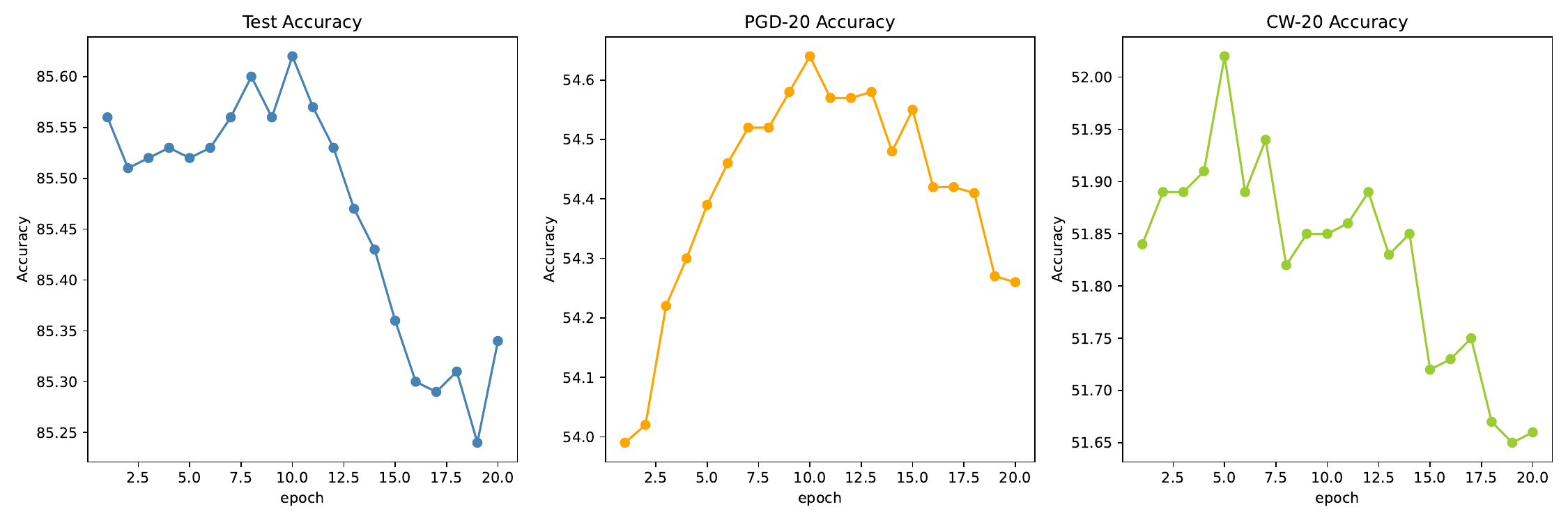}
    \centering
    \caption{The effect of different iterations for HyperAT+}
    \label{fig:HyperAT+ iterations}
\end{figure}

\clearpage
\bigskip

\bibliography{aaai25}

\end{document}